\documentclass{classstyle}
\usepackage{graphicx,amssymb,amsmath,hyperref,booktabs,caption,multirow}
\usepackage{tabularx}
\usepackage{array}

\def\QED{\ensuremath{{\square}}}
\def\markatright#1{\leavevmode\unskip\nobreak\quad\hspace*{\fill}{#1}}

\title{Automated Defect Detection and Grading of Piarom Dates Using Deep Learning}

\author{%
  Nasrin Azimi\thanks{Independent Researcher. Corresponding author. E-mail: Nasrinazimi117@gmail.com}%
  \and
  Danial Mohammad Rezaei\thanks{Department of Computer Engineering, Qazvin Islamic Azad University, Qazvin, Iran}%
}

\FirstAuthor{Azimi~\emph{et.al.}}
\SubTitle{Automated Defect Detection}

\begin{document}
\maketitle\thispagestyle{empty}

\begin{abstract}
Grading and quality control of Piarom dates—a premium and high-value variety cultivated predominantly in Iran—present significant challenges due to the complexity and variability of defects, as well as the absence of specialized automated systems tailored to this fruit. Traditional manual inspection methods are labor-intensive, time-consuming, and prone to human error, while existing AI-based sorting solutions are insufficient for addressing the nuanced characteristics of Piarom dates. In this study, we propose an innovative deep learning framework designed specifically for the real-time detection, classification, and grading of Piarom dates. Leveraging a custom dataset comprising over 9,900 high-resolution images annotated across 11 distinct defect categories, our framework integrates state-of-the-art object detection algorithms and Convolutional Neural Networks (CNNs) to achieve high precision in defect identification. Furthermore, we employ advanced segmentation techniques to estimate the area and weight of each date, thereby optimizing the grading process according to industry standards. Experimental results demonstrate that our system significantly outperforms existing methods in terms of accuracy and computational efficiency, making it highly suitable for industrial applications requiring real-time processing. This work not only provides a robust and scalable solution for automating quality control in the Piarom date industry but also contributes to the broader field of AI-driven food inspection technologies, with potential applications across various agricultural products.
\end{abstract}
\vspace{1cm}

\section{Introduction}
Advancements in artificial intelligence (AI) and deep learning have revolutionized numerous industries, including agriculture, where they have significantly enhanced automation, efficiency, and precision in processes such as planting, harvesting, and quality control~\cite{Kamilaris2018DeepLearning}. Among agricultural products, dates hold substantial economic and cultural importance in many regions, with Piarom dates being one of the most premium and sought-after varieties due to their unique taste, texture, and nutritional value~\cite{Karimi2019Economic}. Predominantly cultivated in the Hormozgan province of Iran, Piarom dates are characterized by their elongated shape, thin skin, and semi-dry texture, making them particularly susceptible to a range of defects during harvesting, handling, and storage.

The grading and quality assessment of Piarom dates are critical for ensuring market competitiveness and consumer satisfaction. However, traditional manual inspection methods are labor-intensive, subjective, and prone to inconsistencies and human error~\cite{Koirala2019,Mishra2017Review}. These challenges are exacerbated by the complex defect profiles of Piarom dates, which include subtle variations in color, texture, and surface anomalies that are difficult to detect and classify accurately. While existing AI-driven sorting systems have been applied to various fruit and vegetable products, they are generally designed for more common varieties and lack the specificity required to address the unique characteristics of Piarom dates.

Recognizing the limitations of current technologies, this study aims to develop an advanced deep learning framework tailored specifically for the real-time detection, classification, and grading of Piarom dates. Our approach involves the creation of a comprehensive and meticulously annotated dataset comprising over 9,900 high-resolution images covering 11 distinct defect categories. To enhance model performance, we employ on-the-fly data augmentation during training, which increases data diversity without expanding the stored dataset size. By integrating state-of-the-art object detection algorithms and Convolutional Neural Networks (CNNs), we seek to achieve high precision in identifying and classifying defects. Additionally, we employ advanced image segmentation techniques to estimate the area and weight of each date, facilitating an optimized grading process that aligns with industry standards.

The proposed system is designed to operate efficiently in real-time industrial settings, addressing the critical need for scalable and accurate automated solutions in the Piarom date industry. Through extensive experiments and evaluations, we demonstrate that our framework significantly outperforms existing methods in terms of accuracy, speed, and robustness under various conditions. This work not only provides a practical solution for enhancing quality control in Piarom date production but also contributes to the broader field of AI-driven agricultural technologies, offering insights and methodologies that can be adapted to other high-value crops with similar challenges.

\section{Related Work}
Over the past decades, the application of technology in agriculture has evolved considerably, particularly in the area of quality assessment and defect detection of agricultural products. Various methods have been explored, ranging from traditional sensor-based approaches to advanced deep learning techniques. In this section, we review the existing literature relevant to defect detection and grading in dates and similar agricultural products, highlighting the progress and limitations of previous studies.

\subsection{Sensor-Based Methods}
Early attempts at automating the grading process of agricultural products relied heavily on sensor-based systems that measured physical attributes such as size, weight, color, and moisture content. These systems employed a range of sensors, including optical, infrared, and ultrasonic devices, to capture measurable properties of the produce. For example, Ismail and Al-Gaadi~\cite{AlGaadi2019FuzzySorting} developed a moisture sensor-based system capable of classifying dates based on moisture levels, which is a key indicator of ripeness and quality. While such systems provided objective measurements and reduced reliance on human inspectors, they were limited in their ability to detect surface defects or internal spoilage that do not significantly affect the measured physical properties.

Advancements in sensor technology led to more sophisticated systems that combined multiple sensor modalities. Al-Gaadi and Ismail~\cite{DateFruitsGradingAlgorithm} introduced a fuzzy logic-based sorting machine that integrated color and ultrasonic sensors to evaluate ripeness and size of dates, achieving improved automation and accuracy. However, these systems often struggled with the detection of subtle defects and required complex calibration and maintenance, limiting their practicality and scalability in industrial applications.

\subsection{Classical Computer Vision Methods}
With the development of computer vision technologies, researchers began exploring image-based analysis for defect detection and grading. Classical computer vision methods involve the extraction of hand-crafted features such as color histograms, texture descriptors, and shape parameters, which are then used for classification using traditional machine learning algorithms.

Al-Ohali~\cite{OldCVALOHALI2011} developed a machine vision system for grading dates by analyzing external characteristics captured through digital imaging. The system extracted features like size, color, and texture to classify dates into different quality levels. While this approach demonstrated the potential of computer vision in automating the grading process, it was limited by the reliance on manually engineered features, which may not capture the full complexity of the defect variations in Piarom dates.

Further studies attempted to improve classification accuracy by integrating additional features and more sophisticated algorithms. For instance, Chen et al.~\cite{Chen2020AppleDetection} proposed an algorithm that combined color and shape features to enhance the grading of fruits. However, classical methods often lack robustness against variations in lighting conditions, orientations, and other environmental factors, and they struggle to generalize across different varieties and defect types.

\subsection{Deep Learning Methods}
The emergence of deep learning has revolutionized the field of computer vision, offering powerful tools for automatic feature extraction and classification. Convolutional Neural Networks (CNNs), in particular, have shown remarkable performance in image recognition tasks, including defect detection in agricultural products.

Lee et al.~\cite{Lee2008} applied CNNs combined with near-infrared imaging to detect defects in apples, achieving high accuracy in identifying surface bruises and blemishes. In the context of date fruits, Al-Qurashi and Al-Ghamdi~\cite{DateYOLOv8} utilized the YOLO (You Only Look Once) object detection framework to classify different varieties of dates and assess ripeness stages, demonstrating the potential for real-time applications.

Despite these advancements, relatively few studies have focused specifically on Piarom dates or addressed the unique challenges associated with their complex defect profiles. The high variability in defect types, coupled with subtle differences in appearance, necessitates a specialized approach. Moreover, existing datasets are often limited in size and diversity, hindering the development of robust models capable of generalizing to real-world conditions.

Our work aims to fill this gap by developing a deep learning framework tailored to the specific requirements of Piarom date grading. By creating a large and diverse dataset with detailed annotations and employing advanced CNN architectures, we seek to overcome the limitations of previous methods and provide a practical solution for the industry.

\section{Dataset}

In this section, we present our meticulously curated custom dataset, designed to facilitate the detection and classification of defects in Piarom dates. Given the unique challenges associated with this premium variety—particularly the variability and complexity of defects—the dataset was crafted to ensure comprehensive coverage of real-world scenarios.

\subsection{Data Generation}

To construct a robust dataset capable of supporting accurate detection and classification of Piarom date defects, we established a controlled image acquisition protocol. The experimental setup was meticulously designed to ensure uniformity and eliminate potential sources of noise that could interfere with subsequent analyses.

\begin{figure}[h]
    \centering
    \includegraphics[width=0.4\textwidth]{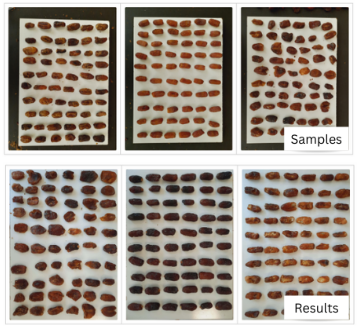}
    \caption{Examples of Piarom date images before and after preprocessing. The top row (\textit{Samples}) displays raw images captured during data collection, with the black background clearly visible. The bottom row (\textit{Results}) presents the images after cropping, where the black background has been removed.}
    \label{fig:piarom_dates}
\end{figure}

We set a black background as the bottom layer of our setup. On top of the black background, we placed a white layer measuring $32\,\text{cm} \times 45\,\text{cm}$, creating a clear separation between the two colors. This configuration facilitated the precise segmentation of dates during preprocessing by providing a stark contrast between the dates and the background. Figure~\ref{fig:piarom_dates} illustrates examples of raw images captured during data collection and the corresponding images after preprocessing.

For each of the 11 defect categories, we captured 18 high-resolution images, each containing 50 dates arranged in a $5 \times 10$ grid, resulting in 198 images and a total of 9,900 date samples.

\subsection{Preprocessing and Data Annotation}

Following image acquisition, the dataset underwent several preprocessing steps to refine the raw data for optimal use in model training. Each high-resolution image was carefully cropped to exclude extraneous background elements, isolating the dates while preserving their real-world dimensions—a crucial factor for size-based classification.

Annotations were performed using the Roboflow platform~\cite{roboflow2023}, adhering to the YOLO format, which includes class labels and bounding box coordinates. This meticulous annotation process generated ground truth data essential for object detection and classification tasks.

\begin{figure}[ht]
    \centering
    \includegraphics[width=0.5\textwidth]{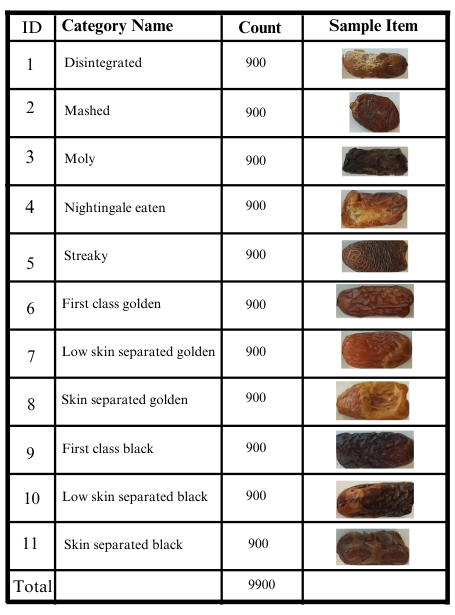}
    \caption{Overview of defect categories in the dataset. Each category includes sample images and the number of images per class (approximately 900), totaling 9,900 samples before augmentation.}
    \label{fig:dataset-classes}
\end{figure}

An overview of the defect categories and their distribution within the dataset is presented in Figure~\ref{fig:dataset-classes}. Each category includes sample images and the number of images per class, ensuring a balanced dataset for training and evaluation.

\subsection{Data Augmentation Techniques}

To enhance the performance of the deep learning models and improve their generalization capabilities, we applied a comprehensive data augmentation strategy implemented during the training process. Data augmentation artificially increases the diversity of the training data without the need to collect new images or expand the stored dataset size, thereby mitigating overfitting and enhancing model robustness~\cite{shorten2019survey}.

 We utilized the Albumentations library~\cite{albumentations2020} for its extensive range of image transformation techniques. The augmentation process included horizontal and vertical flips to create mirror images, rotations (including 90 degrees and upside-down orientations) to account for various positional variances, and adjustments to saturation, brightness, and exposure to simulate different lighting conditions. Specifically, saturation and brightness levels were varied between $-15\%$ and $+15\%$, while exposure adjustments ranged from $-6\%$ to $+6\%$. Additionally, we introduced blur effects with a maximum of 0.6 pixels to mimic out-of-focus scenarios and added noise affecting up to $0.5\%$ of pixels to represent sensor imperfections.

Further refinement of the dataset was achieved through customized YOLO augmentation parameters. These adjustments allowed us to fine-tune the augmentation process to better suit the specific characteristics of Piarom date images. Table~\ref{tab:augmentation} summarizes the default and customized YOLO augmentation parameters utilized in our study.

\begin{table}[h]
\centering
\caption{Customized YOLO augmentation parameters}
\begin{tabularx}{\linewidth}{>{\raggedright\arraybackslash}X
>{\centering\arraybackslash}X
>{\centering\arraybackslash}X}
\toprule
\textbf{Parameter}   & \textbf{Default} & \textbf{Customized} \\
\midrule
hsv\_h        & 0.015                & 0.0                  \\
hsv\_s        & 0.7                  & 0.0                  \\
hsv\_v        & 0.4                  & 0.0                  \\
translate     & 0.1                  & 0.1                  \\
scale         & 0.5                  & 0.1                  \\
fliplr        & 0.5                  & 0.1                  \\
mosaic        & 1.0                  & 0.1                  \\
auto\_augment & randaugment          & custom\_augment      \\
erasing       & 0.4                  & 0.0                  \\
\bottomrule
\end{tabularx}
\label{tab:augmentation}
\end{table}

These augmentation techniques were instrumental in enhancing the diversity of the dataset, thereby improving the generalization capability of the models. By simulating various real-world conditions, the augmented dataset enabled the models to better handle variations in lighting, orientation, and other environmental factors encountered during practical applications.

\subsection{Dataset Composition}

After applying the data augmentation techniques during training, the final dataset comprises over 9,900 labeled images distributed across 11 distinct defect categories, with each category containing approximately 900 images. It is important to note that while data augmentation enriches the diversity of data presented to the model during training, it does not alter the stored dataset size, which remains at 9,900 images. Each category contains approximately 900 images, ensuring a balanced dataset for training and evaluation. To ensure robust model training and evaluation, we performed a train-test split, allocating approximately 78\% of the data for training and 22\% for testing. This split adheres to best practices in machine learning, providing sufficient data for model learning while reserving a representative sample for unbiased performance evaluation~\cite{goodfellow2016deep}.

\begin{figure}[h]
    \centering
    \includegraphics[width=0.47\textwidth]{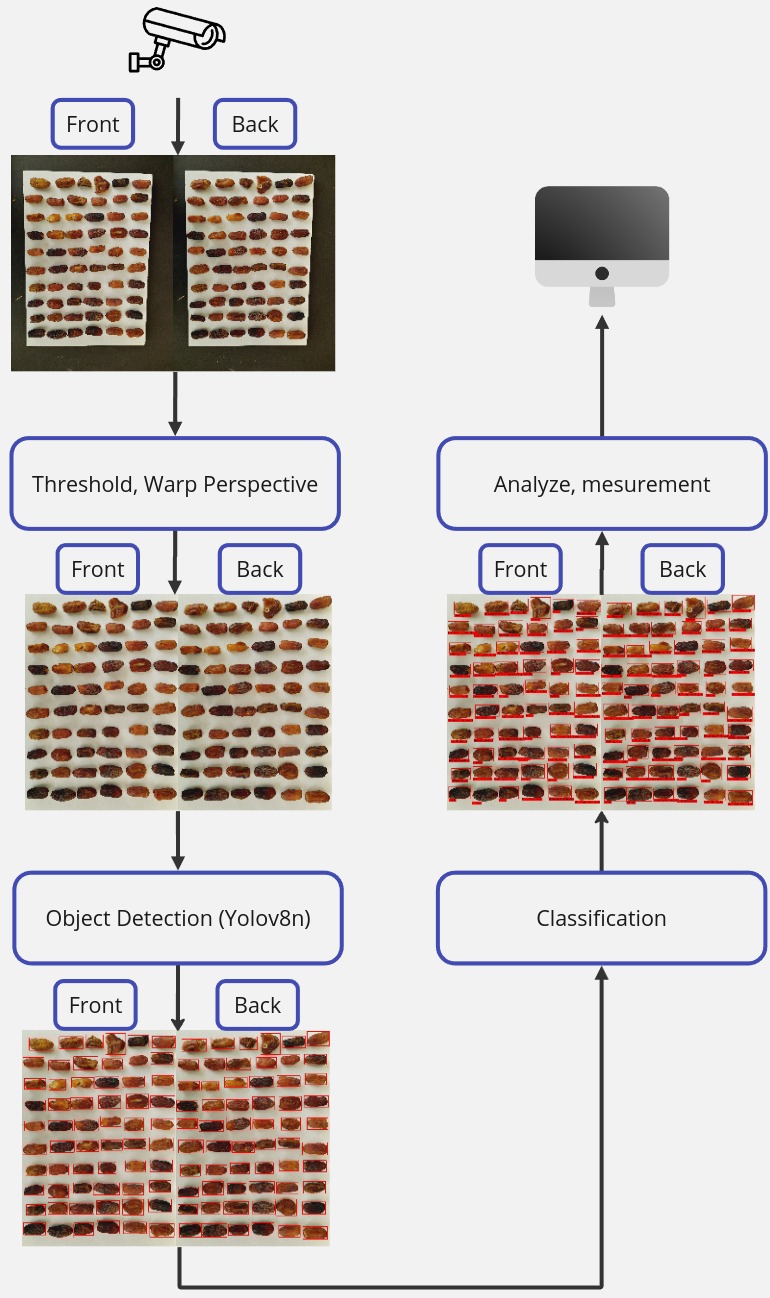}
    \caption{Schematic representation of the Piarom date defect detection and classification process. The flowchart outlines the steps from image acquisition to final analysis.}
    \label{fig:methodology_flow}
\end{figure}

As illustrated in Figure~\ref{fig:methodology_flow}, the dataset creation process began with image acquisition, followed by preprocessing and annotation. These steps formed the foundation of the comprehensive classification dataset. The resulting dataset is meticulously organized and enriched through advanced data augmentation techniques, playing an essential role in training models that deliver high accuracy and efficiency in the automated grading of Piarom dates~\cite{liu2019fruit}.

\subsection{Input and Output Visualization}

\begin{figure}[h]
    \centering
    \includegraphics[width=0.47\textwidth]{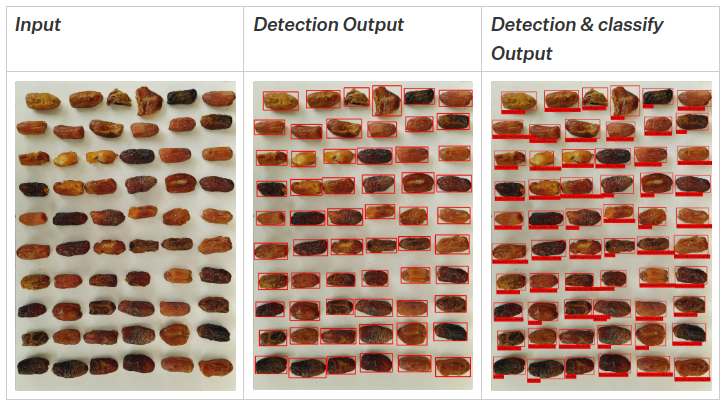}
    \caption{Visualization of the processing pipeline. From left to right: (1) Input image after thresholding and warping, (2) Object detection output with bounding boxes, (3) Detection and classification output with bounding boxes labeled by defect class.}
    \label{fig:input-output}
\end{figure}

To illustrate the transformation of images through the processing pipeline, Figure~\ref{fig:input-output} presents a side-by-side comparison of the input images and the corresponding outputs after object detection and classification. The first image shows the input after thresholding and warping, the second image displays the detection output with visible bounding boxes, and the third image presents the detection and classification output with bounding boxes labeled with the defect class names.

\section{Methodology}
In this section, we present the comprehensive framework employed for the detection and classification of defects in Piarom dates. As illustrated in Figure \ref{fig:methodology_flow}, the methodology encompasses several critical stages: image acquisition, preprocessing, object detection, bounding box alignment, and defect classification. Each stage has been meticulously designed to optimize both the accuracy and computational efficiency of the system, leveraging advanced deep learning models to ensure robust and scalable performance~\cite{Redmon2018}.

\subsection{Image Acquisition}
High-resolution images of Piarom dates were captured in a controlled environment to minimize variability and ensure uniform lighting conditions. The dates were systematically arranged in a $5 \times 10$ grid within a $32 \times 45$ cm box. To ensure comprehensive defect detection, two perspectives were captured for each sample: one from the top view and another from the bottom view. This dual-perspective approach allows for the identification of defects that may not be visible from a single vantage point, thus improving detection accuracy.

\subsection{Preprocessing and Contour Detection}
Upon acquisition, the images underwent a series of preprocessing steps to facilitate accurate detection and classification. The images were first cropped to remove extraneous background elements while preserving the real-world dimensions of the dates. A thresholding technique was then employed to detect the contours of the white box containing the dates. The four vertices of this rectangular contour were identified, allowing for the application of a warp perspective transformation. This step was crucial to aligning the images with real-world dimensions, ensuring consistent scale and orientation for subsequent defect detection.

An example of the preprocessing steps is illustrated in Figure~\ref{fig:input-output}(a). The image shows the preprocessed input after thresholding and warping, which standardizes the image for further processing.

\subsection{Object Detection using YOLO}

For object detection, the YOLO (You Only Look Once) deep learning model was utilized. YOLO's architecture is highly suited for real-time applications as it predicts bounding boxes and class probabilities in a single forward pass through the network~\cite{Redmon2018}. Specifically, YOLO divides the input image into an $S \times S$ grid, with each grid cell predicting $B$ bounding boxes, associated confidence scores, and class probabilities. The object detection process is governed by the following loss function:

\begin{equation}
\begin{aligned}
\mathcal{L} = &\ \lambda_{\text{coord}} \sum_{i=1}^{S^2} \mathbf{1}_i^{\text{obj}} \left[ (x_i - \hat{x}_i)^2 + (y_i - \hat{y}_i)^2 \right. \\
&\ \left. +\ (\sqrt{w_i} - \sqrt{\hat{w}_i})^2 + (\sqrt{h_i} - \sqrt{\hat{h}_i})^2 \right] \\
& +\ \sum_{i=1}^{S^2} \mathbf{1}_i^{\text{obj}} \left[ (C_i - \hat{C}_i)^2 + \sum_{c} (p_i(c) - \hat{p}_i(c))^2 \right] \\
& +\ \lambda_{\text{noobj}} \sum_{i=1}^{S^2} \mathbf{1}_i^{\text{noobj}} (C_i - \hat{C}_i)^2
\end{aligned}
\label{eq:loss_function}
\end{equation}

Where
\begin{itemize}
\item $(x_i, y_i)$ are the predicted coordinates of the bounding box center.
\item $(w_i, h_i)$ represent the width and height of the bounding box.
\item $C_i$ is the confidence score for object presence in the grid cell.
\item $p_i(c)$ is the class probability for class $c$.
\item $\lambda_{\text{coord}}$ and $\lambda_{\text{noobj}}$ control the relative importance of localization and confidence losses, respectively.
\item $\mathbf{1}_i^{\text{obj}}$ is an indicator function that equals 1 if an object is present in cell $i$, and 0 otherwise.
\item $\mathbf{1}_i^{\text{noobj}}$ is an indicator function that equals 1 if no object is present in cell $i$, and 0 otherwise.
\end{itemize}

In the YOLO model, the Intersection over Union (IoU) metric is pivotal for evaluating the accuracy of predicted bounding boxes against the ground truth. The IoU between a predicted bounding box $B_p$ and the ground truth bounding box $B_{gt}$ is defined as:

\begin{equation}
\text{IoU} = \frac{\lvert B_p \cap B_{gt} \rvert}{\lvert B_p \cup B_{gt} \rvert}
\label{eq:iou}
\end{equation}

Where
\begin{itemize}
    \item $\lvert B_p \cap B_{gt} \rvert$ denotes the area of overlap between the predicted bounding box and the ground truth bounding box.
    \item $\lvert B_p \cup B_{gt} \rvert$ represents the total area covered by both the predicted and ground truth bounding boxes.
\end{itemize}

The IoU metric quantifies the degree of overlap between the predicted and actual bounding boxes, providing a measure of localization accuracy. In the context of the YOLO loss function (Equation~\ref{eq:loss_function}), the IoU is incorporated into the confidence score $C_i$, influencing the objectness prediction and the overall loss:

\begin{equation}
C_i = \mathbf{1}_i^{\text{obj}} \times \text{IoU}(B_p, B_{gt})
\label{eq:confidence_score}
\end{equation}

This equation implies that the confidence score $C_i$ is set to the IoU value when an object is present in grid cell $i$, effectively penalizing predictions with low overlap during training. During inference, IoU thresholds are employed to determine true positives and false positives, directly impacting precision and recall metrics.

In this study, the YOLOv8 model~\cite{Jocher2023YOLOv8} was selected due to its balance between accuracy and computational efficiency. Input images were resized to $1280 \times 1280$ pixels to align with the model’s input requirements. Once bounding boxes were generated for each date, a sorting algorithm was applied to align the bounding boxes from the top and bottom perspectives for further analysis.

The output of the object detection stage is depicted in Figure~\ref{fig:input-output}(b), where the detected dates are enclosed within bounding boxes.

\subsection{Bounding Box Sorting and Alignment}
Post-detection, the bounding boxes from the top and bottom views were sorted and aligned to ensure accurate correspondence between the two perspectives. The sorting algorithm first ordered bounding boxes based on their Y-axis coordinates, followed by X-axis sorting within each row. This alignment process was critical for the subsequent comprehensive analysis of defects, allowing for a complete view of each date.

\subsection{Defect Classification}
Once the bounding boxes were aligned, the system applied a defect classification model to detect and categorize defects in each date. We employed a suite of deep learning models, including MobileNetV2~\cite{mobilenetv2}, MobileNetV3~\cite{mobilenetv3}, and YOLOv8-classification~\cite{Jocher2023YOLOv8}, trained on a dataset of 9,900 high-resolution images. To improve the models' generalization capabilities, we employed on-the-fly data augmentation techniques during training. These models were designed to identify defects such as disintegration, mashing, mold formation, and skin separation. The classification process considered both top and bottom views to maximize accuracy, ensuring a thorough inspection of each date.

The final classification results are shown in Figure~\ref{fig:input-output}(c), where each date is labeled with the identified defect class.

\subsection{Output and Analysis}
The final stage of the system generates detailed output reports for each batch of dates, providing key metrics such as:
\begin{itemize}
\item The total number of defects detected.
\item Classification of first-grade dates by color (black and golden).
\item Mean area and weight of the dates.
\end{itemize}
These reports offer valuable insights into the overall quality of the batch, facilitating more informed decision-making during the sorting and grading process.

\section{Experiment Setup}

This section details the experimental setup used for object detection, classification, and defect detection in Piarom dates. All experiments were performed on a high-performance computing system optimized for deep learning tasks, real-time inference, and large-scale dataset processing. The following subsections provide a detailed account of the system configuration, model selection, and performance evaluation.

\subsection{System Configuration}
\label{sec:sys_config}

All experiments, including object detection, real-time classification, and defect detection, were conducted on a system specifically selected to handle the demands of deep learning algorithms. The hardware configuration is presented in Table~\ref{tab:system_config}, with specifications tailored to support computationally intensive tasks, ensuring efficient model training and real-time inference.

\begin{table}[ht]
\centering
\caption{System Configuration Details}
\begin{tabularx}{\linewidth}{Xr}
\toprule
\textbf{Component}   & \textbf{Specification} \\
\midrule
Operating System     & Ubuntu 22.04 LTS (64-bit)        \\
Processor            & Intel Core™ i7-11800H            \\
Cores                & 8 cores, 16 threads              \\
Base Clock Speed     & 2.3 GHz (up to 4.6 GHz)          \\
Memory               & 32.0 GiB DDR4 RAM                \\
GPU                  & NVIDIA GeForce RTX™ 3060         \\
VRAM                 & 6144 MiB (6 GB)                  \\
Storage              & 1 TB NVMe Solid-State Drive (SSD)\\
\bottomrule
\end{tabularx}
\label{tab:system_config}
\end{table}

\begin{figure}[h]
\centering
\includegraphics[width=0.45\textwidth]{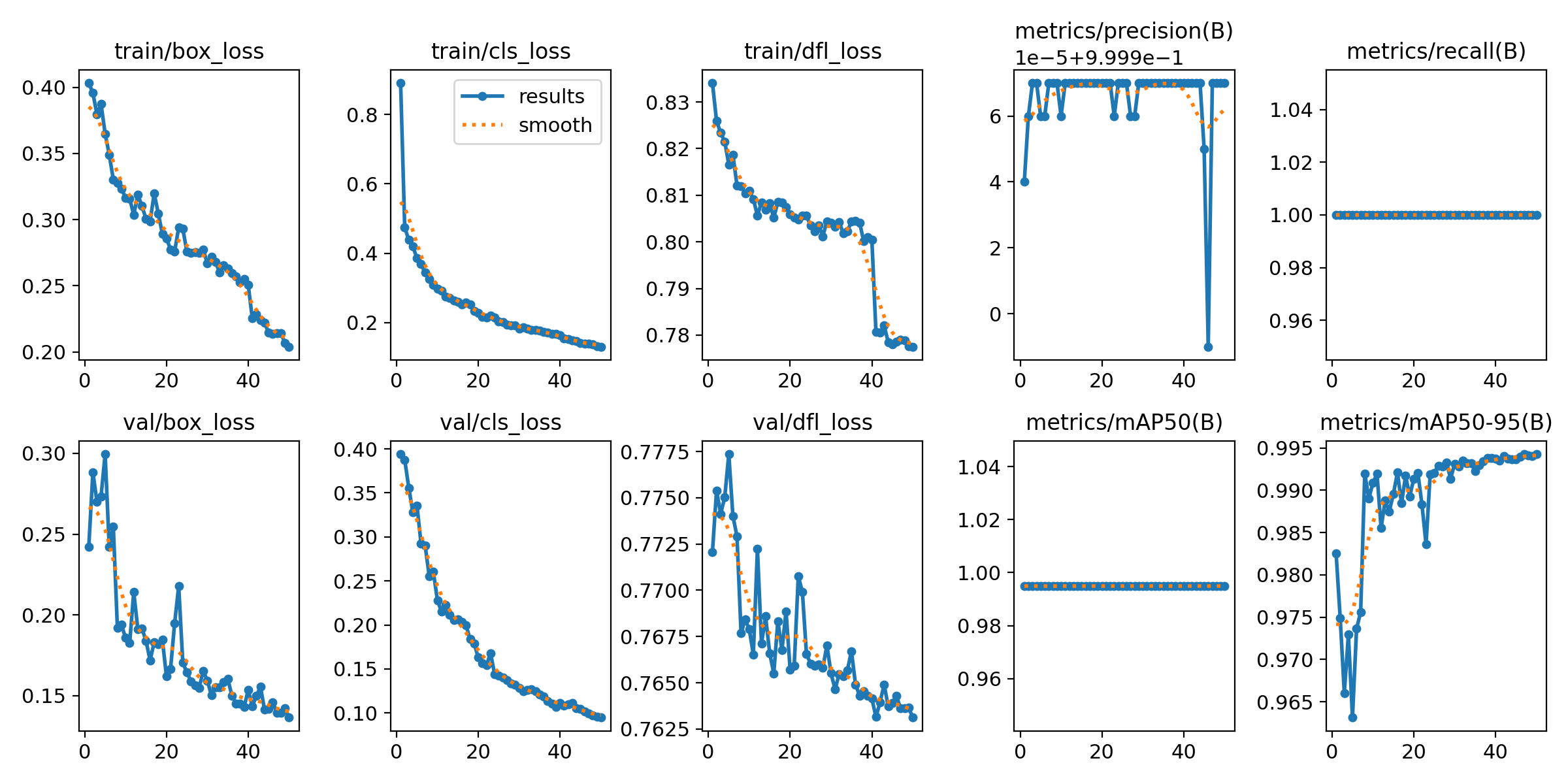}
\caption{Training and validation loss curves for YOLOv8-nano.}
\label{fig:nano_losses}
\end{figure}

\begin{figure}[h]
\centering
\includegraphics[width=0.45\textwidth]{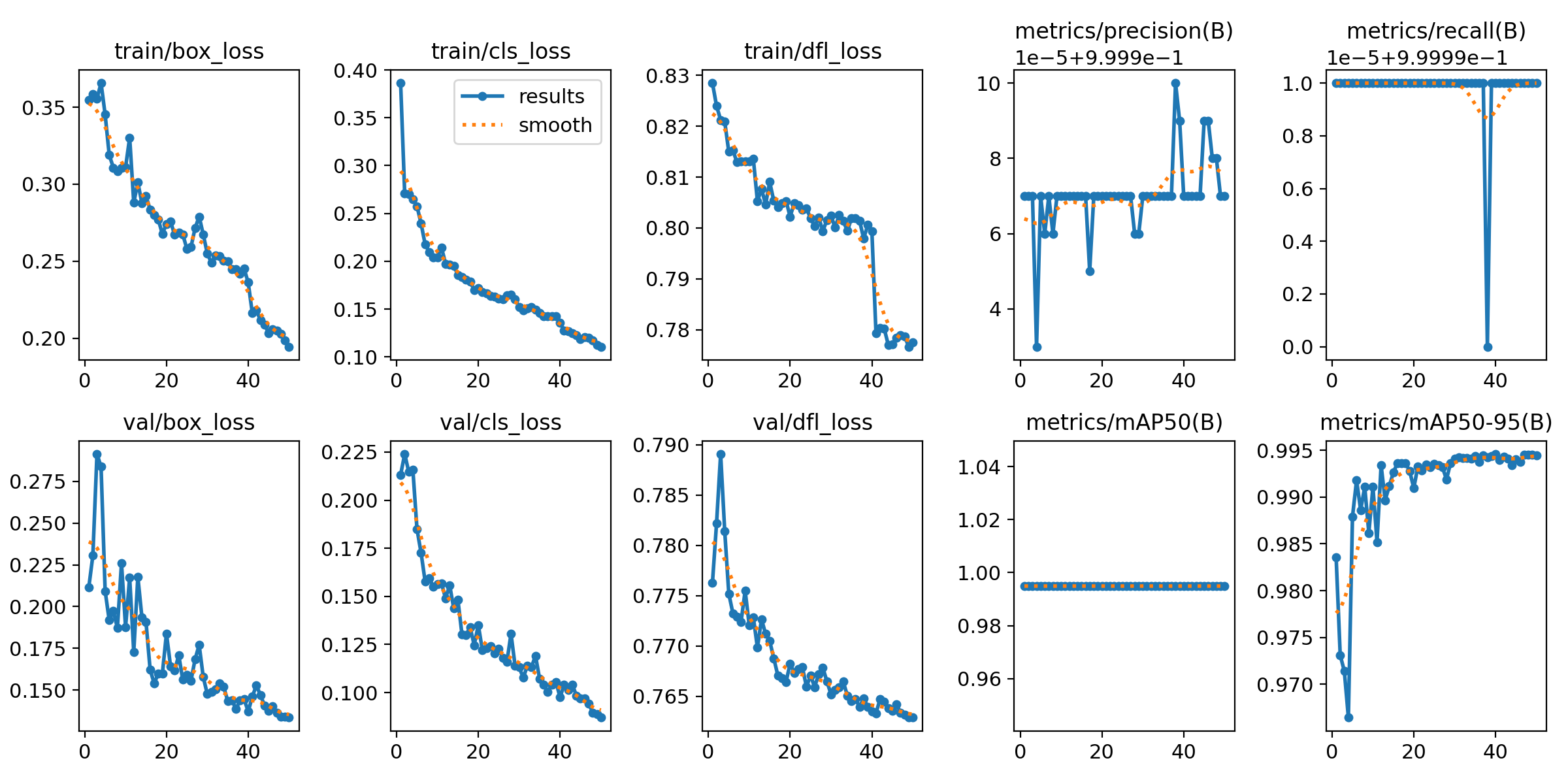}
\caption{Training and validation loss curves for YOLOv8-small.}
\label{fig:small_losses}
\end{figure}

\begin{figure}[h]
\centering
\includegraphics[width=0.45\textwidth]{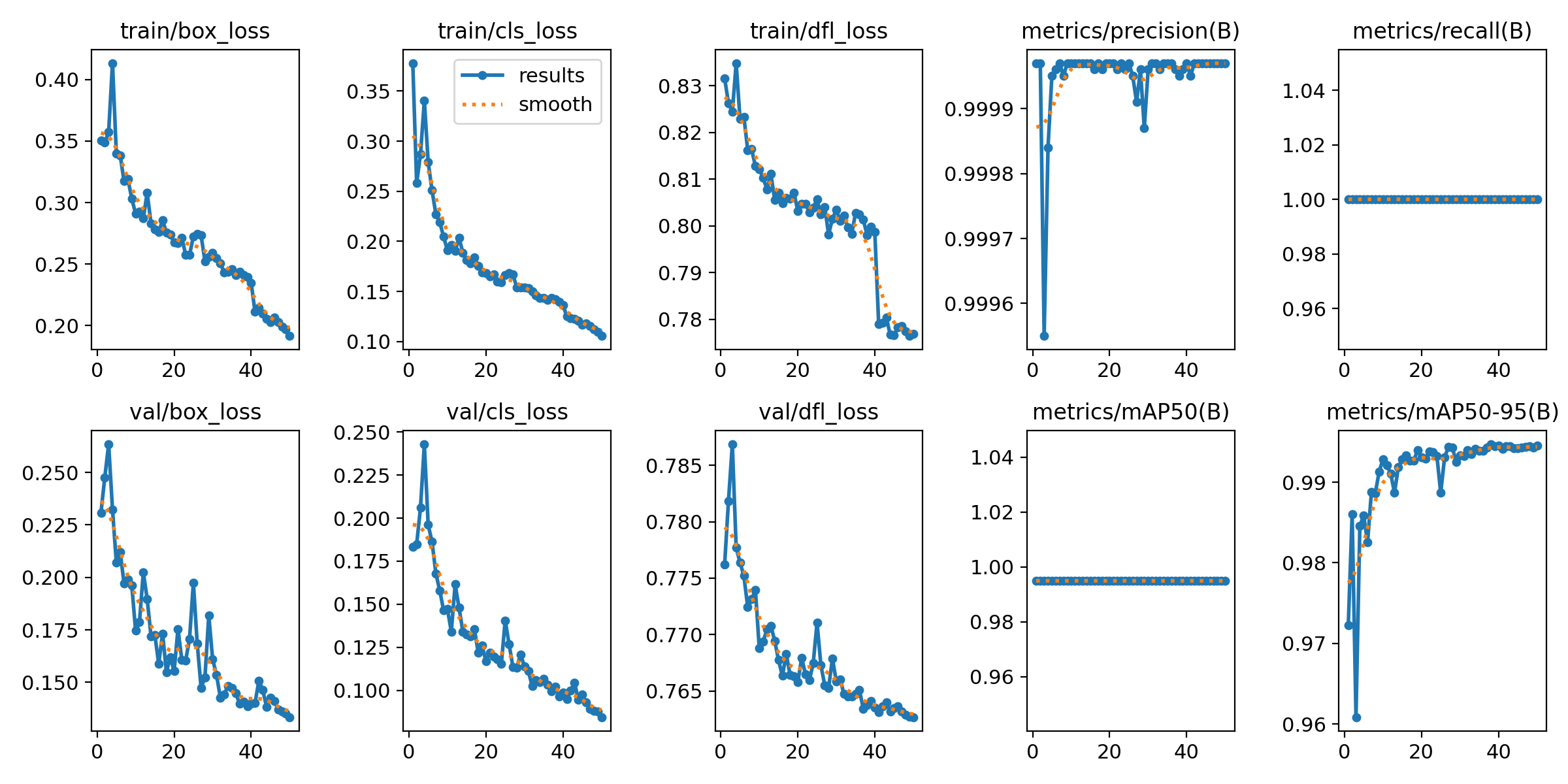}
\caption{Training and validation loss curves for YOLOv8-medium.}
\label{fig:medium_losses}
\end{figure}

\begin{figure}[ht]
\centering
\includegraphics[width=0.5\textwidth]{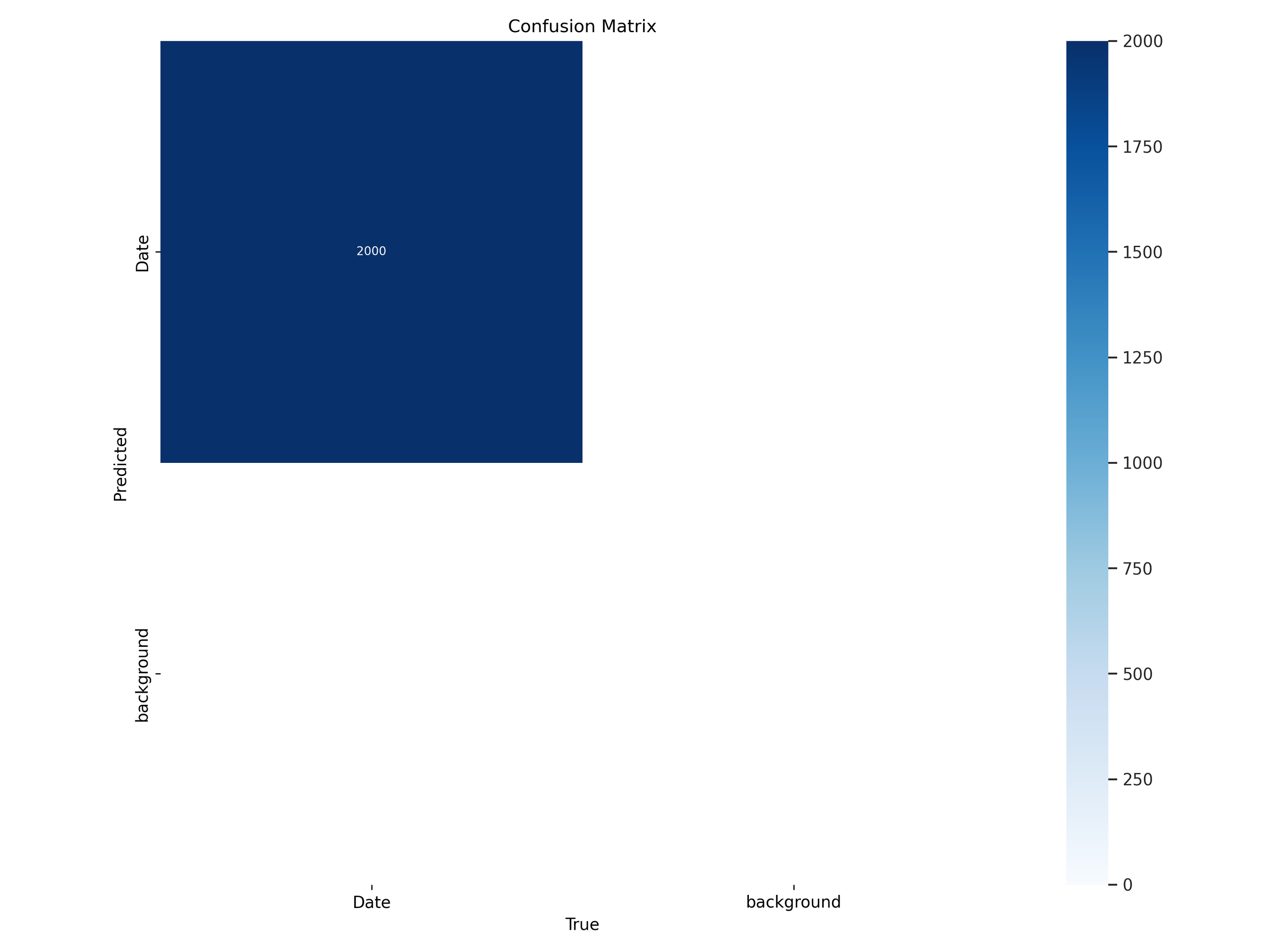}
\caption{YOLOv8n Object Detection Confusion Matrix}
\label{fig:nano_confusion}
\end{figure}

\begin{figure}[ht]
\centering
\includegraphics[width=0.5\textwidth]{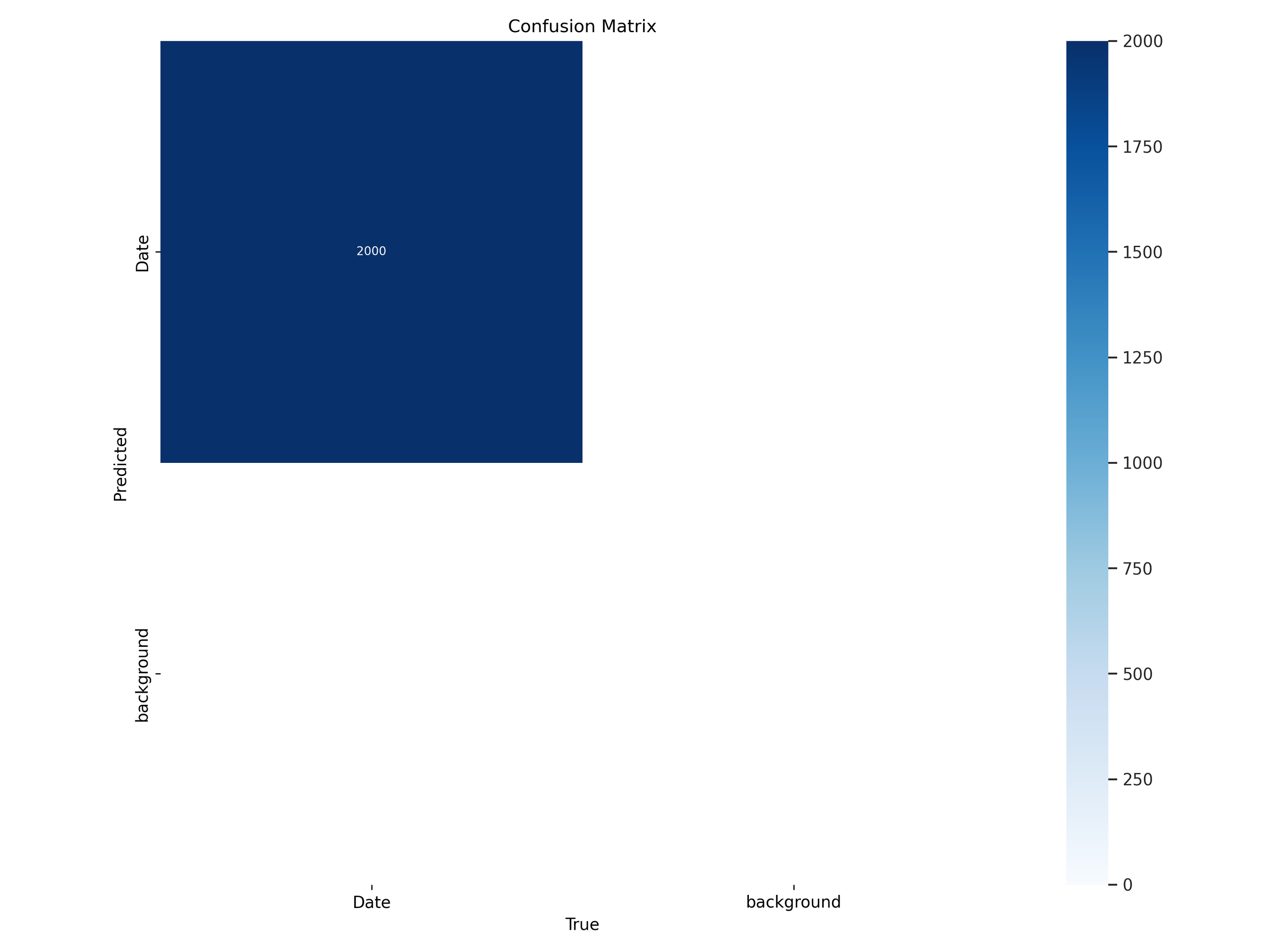}
\caption{YOLOv8s Object Detection Confusion Matrix}
\label{fig:small_confusion}
\end{figure}

\begin{figure}[ht]
\centering
\includegraphics[width=0.5\textwidth]{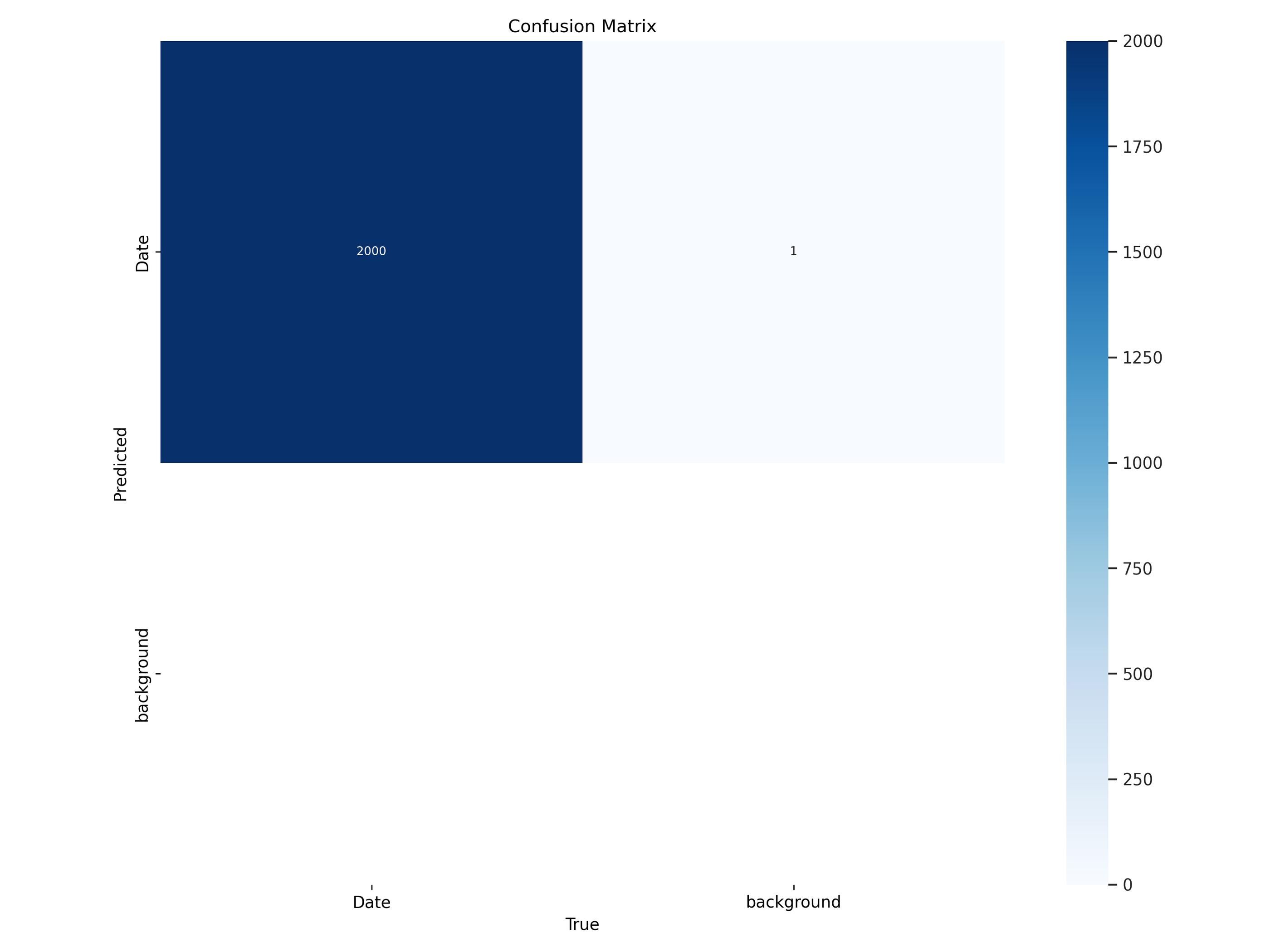}
\caption{YOLOv8m Object Detection Confusion Matrix}
\label{fig:medium_confusion}
\end{figure}

The system’s configuration, as detailed in Table~\ref{tab:system_config}, was instrumental in ensuring rapid inference times and efficient resource usage during all stages of model training and evaluation. The combination of a multi-core processor and a high-performance GPU, along with ample memory, provided the necessary computational power to handle large datasets and high-resolution images, facilitating the efficient execution of deep learning algorithms.

\begin{figure}[ht]
\centering
\includegraphics[width=0.5\textwidth]{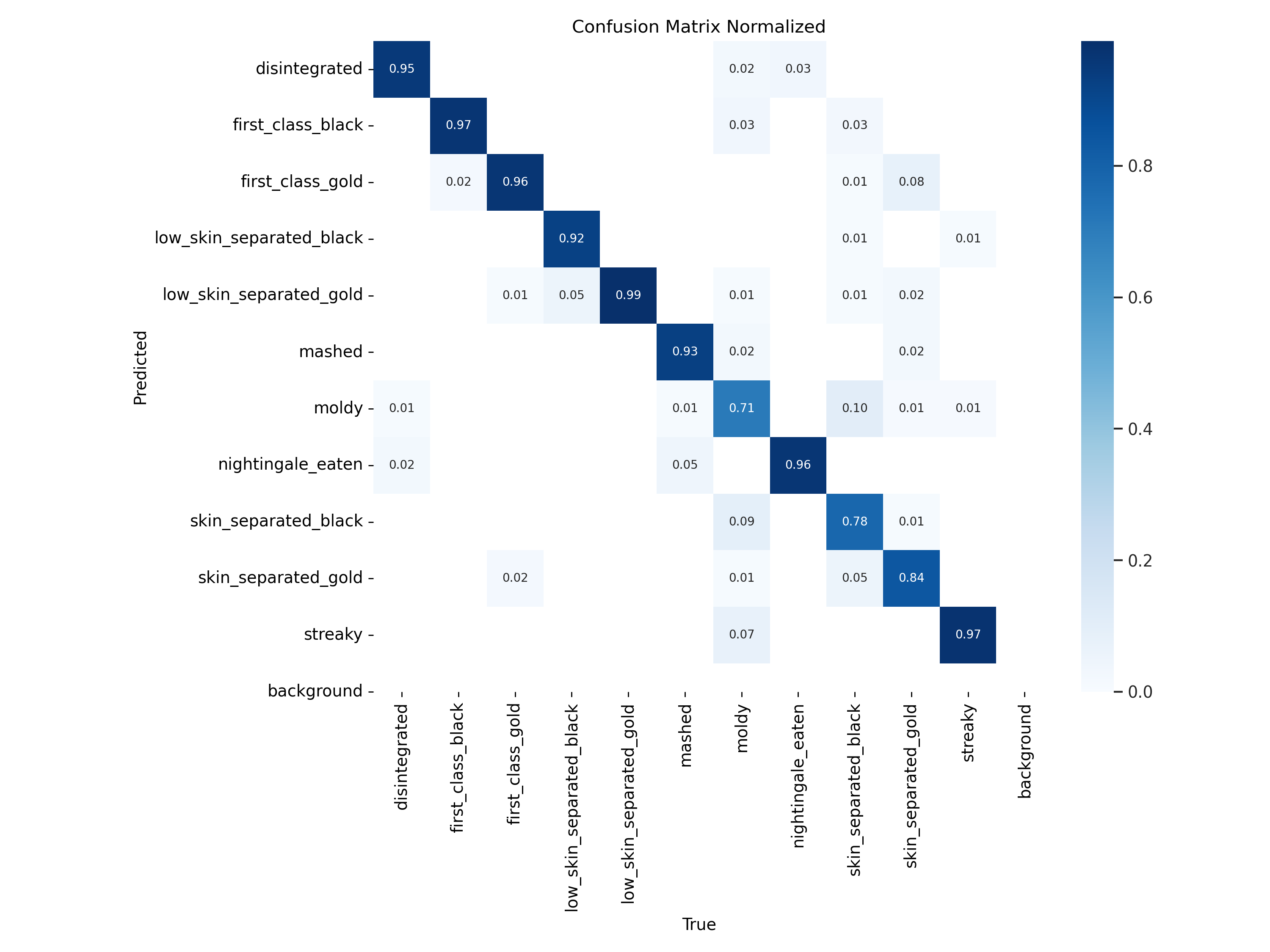}
\caption{YOLOv8n Classification Confusion Matrix}
\label{fig:class-nano_confusion}
\end{figure}

\begin{figure}[ht]
\centering
\includegraphics[width=0.5\textwidth]{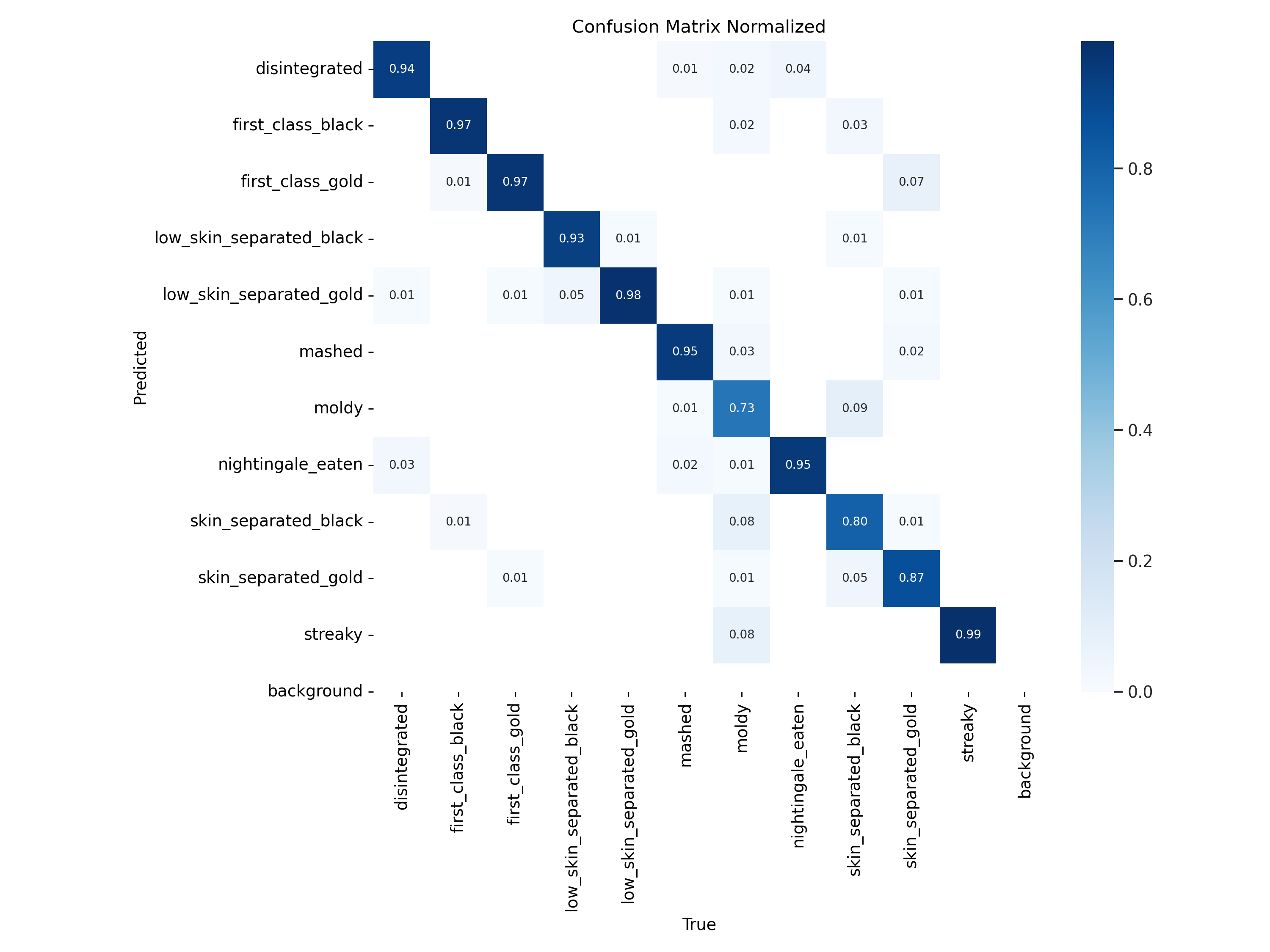}
\caption{YOLOv8s Classification Confusion Matrix}
\label{fig:class-small_confusion}
\end{figure}

\begin{figure}[ht]
\centering
\includegraphics[width=0.5\textwidth]{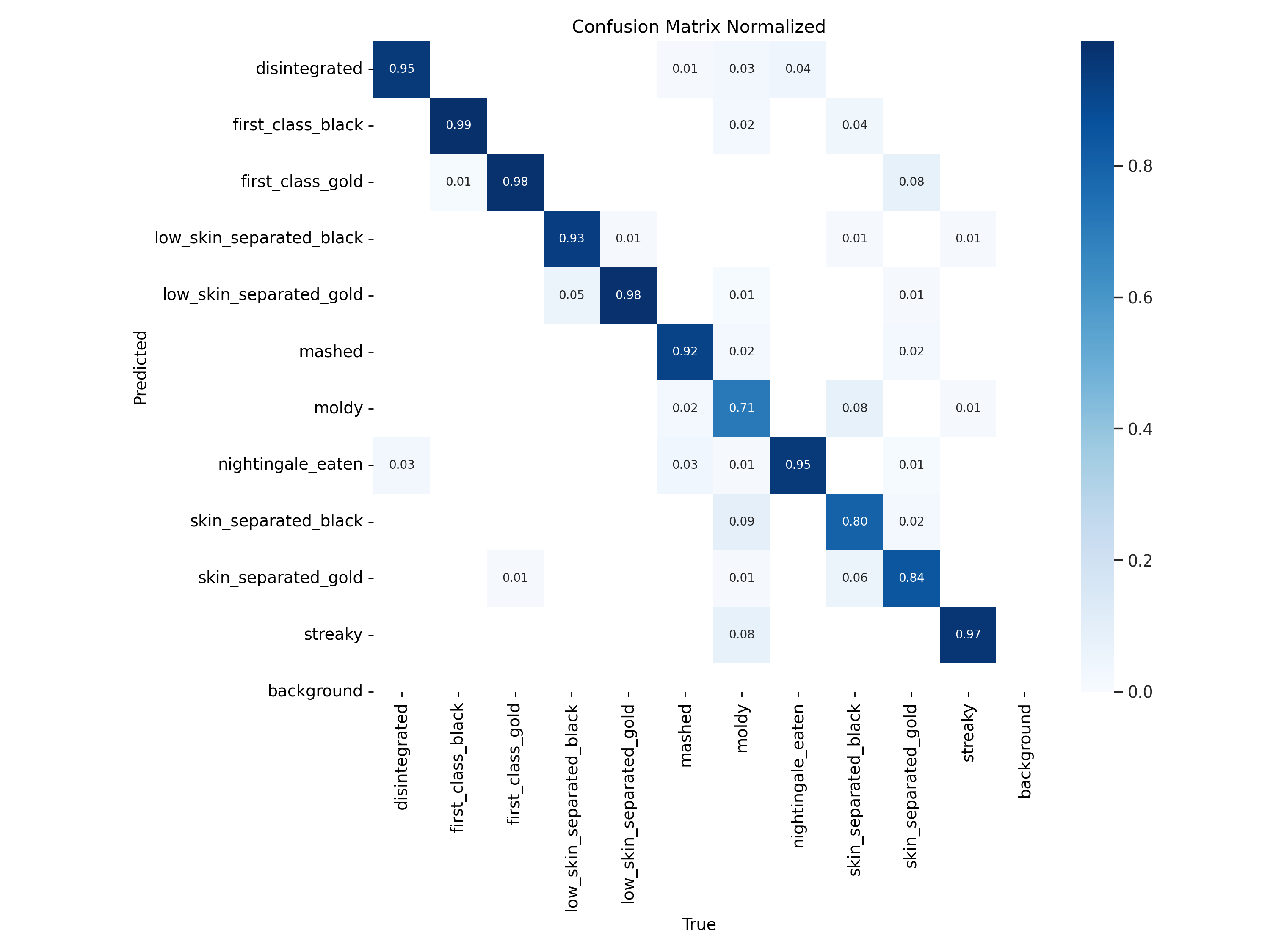}
\caption{YOLOv8m Classification Confusion Matrix}
\label{fig:class-medium_confusion}
\end{figure}

\subsection{Object Detection Results}

In 2023, YOLOv8~\cite{Jocher2023YOLOv8} models emerged as a benchmark for object detection tasks, recognized globally for their state-of-the-art performance and computational efficiency. For this study, we evaluated three variants of the YOLOv8 series—YOLOv8-nano, YOLOv8-small, and YOLOv8-medium—due to their suitability for large input images (1280x1280 pixels) and real-time processing capabilities.

The training and validation loss curves for YOLOv8 models are shown in Figures~\ref{fig:nano_losses}, \ref{fig:small_losses}, and \ref{fig:medium_losses}. The box loss and classification loss decrease steadily over time, with YOLOv8-medium converging slightly faster than the nano and small models. The validation losses show minimal overfitting, indicating that the models generalize well to unseen data.

The precision, recall, and mAP metrics for YOLOv8 models are summarized in Table~\ref{ObjectDetectionTable}, with confusion matrices provided in Figures~\ref{fig:nano_confusion}, \ref{fig:small_confusion}, and \ref{fig:medium_confusion}. YOLOv8-medium achieved the highest mAP50 score, while YOLOv8-nano demonstrated the fastest inference speed but with a slightly lower mAP50-95 score. These metrics suggest that while YOLOv8-nano is suitable for real-time applications, YOLOv8-medium offers a better balance between accuracy and inference time for applications where higher precision is required.

\begin{table*}[ht]
\centering
\caption{Evaluation of YOLOv8 Models: Performance Metrics and Resource Requirements for Object Detection}
\begin{tabular}{@{}lccccccc@{}}
\toprule
\textbf{Model} & F1-Score & Precision & Recall & mIoU & mAP 0.5-0.95 & Inference Time (ms) & GPU Usage \\
\midrule
\multicolumn{8}{@{}c@{}}{\textbf{Resolution: 1280x1280}} \\
YOLOv8m & 0.99998 & 0.99992 & 1.00000 & 0.97470 & 0.99471 & 61.20000 & 5152 MiB \\
YOLOv8s & 0.99999 & 0.99997 & 1.00000 & 0.98156 & 0.99471 & 34.60000 & 3944 MiB \\
YOLOv8n & 0.99999 & 0.99997 & 1.00000 & 0.98131 & 0.99459 & 12.80000 & 2236 MiB \\
\midrule
\multicolumn{8}{@{}c@{}}{\textbf{Resolution: 640x640}} \\
YOLOv8m & 0.99960 & 0.99920 & 1.00000 & 0.96270 & 0.95630 & 36.00000 & 520 MiB \\
YOLOv8s & 0.99960 & 0.99920 & 1.00000 & 0.96170 & 0.95570 & 9.40000 & 388 MiB \\
YOLOv8n & 0.99960 & 0.99920 & 1.00000 & 0.38610 & 0.95050 & 6.50000 & 324 MiB \\
\bottomrule
\label{ObjectDetectionTable}
\end{tabular}
\end{table*}

\begin{table*}[ht]
\centering
\captionsetup{justification=centering}
\caption{Evaluation of Models: Performance Metrics and Resource Requirements for Classification. All GPU usages are measured out of a total of 6144 MiB of GPU memory.}
\begin{tabular}{@{}l@{\hskip 4pt}c@{\hskip 4pt}c@{\hskip 4pt}c@{\hskip 4pt}c@{\hskip 4pt}c@{\hskip 4pt}c@{}}
\toprule
\textbf{Model} & \textbf{Accuracy} & \textbf{F1-score} & \textbf{Precision} & \textbf{Recall} & \textbf{Inference Time (ms)} & \textbf{GPU Usage (MiB)} \\
\midrule
\multicolumn{7}{@{}c@{}}{\textbf{Resolution: 480x480}} \\
YOLOv8n-cls               & 0.9100       & 0.9081       & 0.9099        & 0.9086     & 0.7     & 613      \\
YOLOv8s-cls               & 0.9195       & 0.9179       & 0.9195        & 0.9185     & 1.7     & 787      \\
YOLOv8m-cls               & 0.9136       & 0.9117       & 0.9136        & 0.9122     & 4.7     & 1047     \\
MobileNetV2              & 0.9027       & 0.9027       & 0.9027        & 0.9027     & 29.7    & 505      \\
MobileNetV3-Large        & 0.9095       & 0.9095       & 0.9095        & 0.9095     & 27.7    & 489      \\
MobileNetV3-Small        & 0.8954       & 0.8954       & 0.8954        & 0.8954     & 27.3    & 277      \\
ResNet18                  & 0.9018       & 0.9018       & 0.9018        & 0.9018     & 25.5    & 405      \\
\bottomrule
\label{Classification}
\end{tabular}
\end{table*}

\subsection{Classification Results}

In addition to object detection, we evaluated several classification models for defect detection, including YOLOv8-classification variants and MobileNet architectures. The performance metrics are summarized in Table~\ref{Classification}. The YOLOv8-small classification model achieved the highest accuracy of 91.95\%, outperforming the MobileNet and ResNet18 models in both accuracy and inference time.

\section{Conclusion}

The experimental results clearly demonstrate the robustness of the YOLOv8 models for object detection and classification tasks. Both the object detection and classification models achieved high accuracy while maintaining efficient resource usage, making them highly suitable for real-time deployment in environments with constrained computational resources. The use of padding over resizing in the classification task further ensured high precision in defect detection, preserving the integrity of the original images and minimizing the risk of accuracy loss. These results position our approach as an effective and scalable solution for the automated detection and classification of defects in Piarom dates, with potential applications in other agricultural sectors~\cite{Mekala2021}.

\section{Limitations}

While the proposed deep learning framework exhibits strong performance in the defect detection and classification of Piarom dates, the study encountered notable challenges related to the visual similarity between certain defect categories. Specifically, four classes---Low Skin Separated Black, Low Skin Separated Golden, Skin Separated Black, and Skin Separated Golden---were identified as being particularly difficult for the model to distinguish due to their close resemblance in appearance. This visual overlap often led to misclassification, which highlights a significant limitation in the system's ability to accurately differentiate between these subtle defect types~\cite{Seyyedhasani2021}. This limitation is further illustrated in the classification confusion matrices of the various YOLO models (Figures~\ref{fig:class-nano_confusion},~\ref{fig:class-small_confusion}, and~\ref{fig:class-medium_confusion}), where the highest rates of misclassification are observed between these specific categories.

Despite increasing the size of the dataset, the close visual proximity of these categories suggests that simply adding more data may not substantially improve the model's discriminatory power. Future research should consider alternative approaches to address this issue, such as refining the model architecture to better handle fine-grained visual distinctions or integrating additional data modalities. For instance, incorporating multimodal data, such as combining visual inputs with spectral or textural analysis, could enhance the system's ability to differentiate between these visually similar categories, thereby improving overall classification accuracy~\cite{Papakostas2020}.

\section*{Data and Code Availability}

The source code and dataset used in this study are publicly available at \url{https://github.com/nasrin117/Piarom-datenet}. The repository includes comprehensive documentation and examples to facilitate replication and further research.

\bibliographystyle{plain}
\bibliography{citations}

\begin{thebibliography}{10}

\bibitem{AlGaadi2019FuzzySorting}
Khalid Al-Gaadi and Hamed Ismail.
\newblock Automated date fruits sorting machine using fuzzy logic controller.
\newblock {\em International Journal of Recent Technology and Engineering}, 8(2S3):658--663, 2019.

\bibitem{DateFruitsGradingAlgorithm}
Khalid Al-Gaadi and Hamed Ismail.
\newblock Date fruits grading and sorting classification algorithm using colors and shape features.
\newblock {\em International Journal of Engineering Research and Technology}, 13(8):1917--1920, 2020.

\bibitem{OldCVALOHALI2011}
Yousef Al-Ohali.
\newblock Computer vision based date fruit grading system: Design and implementation.
\newblock {\em Journal of King Saud University-Computer and Information Sciences}, 23(1):29--36, 2011.

\bibitem{DateYOLOv8}
H.~Al-Qurashi and S.~Al-Ghamdi.
\newblock Date fruit detection and classification based on its variety using deep learning technology.
\newblock {\em IEEE Access}, 11:74567--74578, 2023.

\bibitem{albumentations2020}
Alexander Buslaev, Vladimir~I. Iglovikov, Evgenii Khvedchenya, Artem Parinov, Mikhail Druzhinin, and Alexey~A. Kalinin.
\newblock Albumentations: Fast and flexible image augmentations.
\newblock {\em Information}, 11(2):125, 2020.

\bibitem{Chen2020AppleDetection}
Junyi Chen, Min Hu, and Jianfeng Lu.
\newblock Application of deep learning in apple fruit detection and recognition.
\newblock {\em Pattern Recognition Letters}, 128:120--126, 2020.

\bibitem{roboflow2023}
Brad Dwyer, Joseph Nelson, Tyson Hansen, et~al.
\newblock Roboflow (version 1.0) [software], 2023.
\newblock Computer vision platform.

\bibitem{goodfellow2016deep}
Ian Goodfellow, Yoshua Bengio, and Aaron Courville.
\newblock {\em Deep Learning}.
\newblock MIT Press, 2016.

\bibitem{mobilenetv3}
Andrew Howard, Ruoming Pang, Hartwig Adam, Quoc~V. Le, Bo~Chen, Wei Wang, Liang-Chieh Xu, Mark Chen, Mingxing Tan, Guanhua Chu, Vijay Vasudevan, and Yukun Zhu.
\newblock Searching for mobilenetv3.
\newblock In {\em Proceedings of the IEEE International Conference on Computer Vision}, pages 1314--1324, 2019.

\bibitem{Jocher2023YOLOv8}
Glenn Jocher, Ayush Chaurasia, Jing Qiu, and Adam Stoken.
\newblock {YOLO by Ultralytics}.
\newblock 2023.

\bibitem{Kamilaris2018DeepLearning}
Andreas Kamilaris and Francesc~Xavier Prenafeta-Bold{\'u}.
\newblock Deep learning in agriculture: A survey.
\newblock {\em Computers and Electronics in Agriculture}, 147:70--90, 2018.

\bibitem{Karimi2019Economic}
S.~Karimi and S.~M. Hosseini.
\newblock Economic importance of date production in iran and the world: A review.
\newblock {\em International Journal of Horticultural Science and Technology}, 6(2):259--270, 2019.

\bibitem{Koirala2019}
Arun Koirala, Kerry Walsh, Zhe Wang, and Carol McCarthy.
\newblock Deep learning--method overview and review of use in plant disease detection.
\newblock {\em Computers and Electronics in Agriculture}, 162:219--235, 2019.

\bibitem{Lee2008}
Dah-Jye Lee, Hao Ren, and Yi-Ping~Phoebe Huang.
\newblock A computer vision system for defect grading of date fruits.
\newblock {\em Journal of Food Engineering}, 84(3):430--438, 2008.

\bibitem{liu2019fruit}
Jinhui Liu, Xuyang Wang, and Shujie Yang.
\newblock Fruit detection using improved yolov3 model.
\newblock {\em Precision Agriculture}, 20(4):803--822, 2019.

\bibitem{Mekala2021}
Anudeep Mekala, Prakash Venkatesan, and Jeyaraman Rajendran.
\newblock A novel deep learning framework for defect detection and grading in agricultural products.
\newblock {\em Information Processing in Agriculture}, 8(1):45--54, 2021.

\bibitem{Mishra2017Review}
Pawan Mishra, Mohd Syuhaimi~Mohd Asaari, Ana Herrero-Langreo, Santosh Lohumi, Belen Diezma, and Paul Scheunders.
\newblock Close range hyperspectral imaging of plants: A review.
\newblock {\em Biosystems Engineering}, 164:49--67, 2017.

\bibitem{Papakostas2020}
Michail Papakostas, Apostolos Giannakopoulos, and Christos Stylios.
\newblock Deep learning frameworks for robotic sorting of agricultural products.
\newblock {\em Computers and Electronics in Agriculture}, 174:105421, 2020.

\bibitem{Redmon2018}
Joseph Redmon and Ali Farhadi.
\newblock Yolov3: An incremental improvement.
\newblock arXiv preprint arXiv:1804.02767, 2018.

\bibitem{mobilenetv2}
Mark Sandler, Andrew Howard, Menglong Zhu, Andrey Zhmoginov, and Liang-Chieh Chen.
\newblock Mobilenetv2: Inverted residuals and linear bottlenecks.
\newblock In {\em Proceedings of the IEEE Conference on Computer Vision and Pattern Recognition}, pages 4510--4520, 2018.

\bibitem{Seyyedhasani2021}
Hamed Seyyedhasani, John Noland, William~M. Miller, Cristina~M. Sabliov, Sanjay~C. Negi, and Hongwei Xin.
\newblock Machine learning approaches to improve non-destructive quality assessment of agricultural products: A case study on apples.
\newblock {\em Postharvest Biology and Technology}, 172:111376, 2021.

\bibitem{shorten2019survey}
Connor Shorten and Taghi~M. Khoshgoftaar.
\newblock A survey on image data augmentation for deep learning.
\newblock {\em Journal of Big Data}, 6(1):1--48, 2019.

\end{thebibliography}

\end{document}